\documentclass[letterpaper]{article} 
\usepackage{aaai2026}  
\usepackage{times}  
\usepackage{helvet}  
\usepackage{courier}  
\usepackage[hyphens]{url}  
\usepackage{graphicx} 
\usepackage{natbib}  
\usepackage{caption} 
\usepackage{xcolor}
\usepackage{colortbl}
\usepackage{algorithm}
\usepackage{algorithmic}
\usepackage{amsfonts}
\usepackage{amsmath}
\usepackage{multirow}
\usepackage{booktabs}
\usepackage{bm}

\usepackage{newfloat}
\usepackage{listings}
\DeclareCaptionStyle{ruled}{labelfont=normalfont,labelsep=colon,strut=off} 
\floatstyle{ruled}
\newfloat{listing}{tb}{lst}{}
\floatname{listing}{Listing}
\title{Personality-guided Public-Private Domain Disentangled Hypergraph-Former Network for Multimodal Depression Detection}
\author {
    Changzeng Fu\textsuperscript{\rm 1,}\thanks{Corresponding Author},
    Shiwen Zhao\textsuperscript{\rm 1},
    Yunze Zhang\textsuperscript{\rm 1},
    Zhongquan Jian\textsuperscript{\rm 2},
    Shiqi Zhao\textsuperscript{\rm 1},
    Chaoran Liu\textsuperscript{\rm 3}
}
\affiliations {
    \textsuperscript{\rm 1}School of Computer Science and Engineering, Northeastern University, Shenyang, China\\
    \textsuperscript{\rm 2}School of Computer and Data Science, Minjiang University, Fuzhou, China\\
    \textsuperscript{\rm 3}Research and Development Center for Large Language Models, NII, Tokyo, Japan\\
    
    fuchangzeng@qhd.neu.edu.cn,
    2472505@stu.neu.edu.cn,
    2402087@stu.neu.edu.cn, \\
    jianzq@mju.edu.cn,
    zhaoshiqi@qhd.neu.edu.cn,
    cliu@nii.ac.jp
    
}

\begin{document}

\maketitle

\begin{abstract}
Depression represents a global mental health challenge requiring efficient and reliable automated detection methods. Current Transformer- or Graph Neural Networks (GNNs)-based multimodal depression detection methods face significant challenges in modeling individual differences and cross-modal temporal dependencies across diverse behavioral contexts. Therefore, we propose P$^3$HF (Personality-guided Public-Private Domain Disentangled Hypergraph-Former Network) with three key innovations: (1) personality-guided representation learning using LLMs to transform discrete individual features into contextual descriptions for personalized encoding; (2) Hypergraph-Former architecture modeling high-order cross-modal temporal relationships; (3) event-level domain disentanglement with contrastive learning for improved generalization across behavioral contexts. Experiments on MPDD-Young dataset show P$^3$HF achieves around 10\% improvement on accuracy  and weighted F1 for binary and ternary depression classification task over existing methods. Extensive ablation studies validate the independent contribution of each architectural component, confirming that personality-guided representation learning and high-order hypergraph reasoning are both essential for generating robust, individual-aware depression-related representations. The code is released at https://github.com/hacilab/P3HF.
\end{abstract}

\section{Introduction}

Depression affects approximately 3.8\% of the global population and remains the fourth leading cause of death among individuals aged 15-29, with over 700,000 suicide deaths annually~\cite{WHO}. The severe shortage of mental healthcare resources, particularly in developing regions, has driven the development of automated depression detection technologies for early diagnosis~\cite{trotzek2018utilizing,niu2024pointtransform,fu2025m,zhao2025chinese,fu2025himul}.

The methodological development of automated depression detection has progressively transcended unimodal paradigms toward richer multimodal representations. Early recurrent neural networks~\cite{ma2016depaudionet} established temporal sequence modeling; these were superseded by Transformer-based contextual encoders~\cite{meng2021bidirectional} that capture long-range dependencies, and further refined through graph convolutional networks~\cite{qin2022using} and hypergraph-based structural modeling~\cite{li2024hypergraph}, enabling nuanced extraction of cross-modal semantic relations and spatio-temporal dynamics. Complementing these architectural advances, the field has shifted from population-level models to individualized perspectives, explicitly incorporating personality factors as moderators of depression severity~\cite{francis2023personality, fu2025facial}. These methodological maturation is mirrored by dataset evolution: beginning with the interview-centric, single-event AVEC-2014 and DAIC-WOZ corpora, progressing to the extended E-DAIC, and culminating in the recently introduced MPDD benchmark that uniquely integrates individual-difference annotations with multi-event multimodal recordings to enable personalized detection.

Despite these advances, three critical problems remain unresolved. \textbf{Problem 1:} depression manifestations exhibit significant individual variations, resulting in highly heterogeneous multimodal features. Existing deep learning methods predominantly adopt uniform modeling strategies, failing to account for individual differences in expression patterns and communication styles. \textbf{Problem 2:} while hypergraph neural networks effectively model high-order cross-modal relationships through hyperedges connecting multiple nodes, these hyperedges constitute unordered sets that cannot explicitly capture temporal sequential relationships between nodes. This limitation is particularly problematic for depression detection, where symptom expression often exhibits crucial temporal dependencies. \textbf{Problem 3:} depression as a complex psychological disorder demonstrates significant context-dependent manifestations that single-event scenarios cannot comprehensively capture.

According to Bandura's reciprocal determinism theory, individual depression manifestations result from mutual influences among personal factors, environment, and behavior~\cite{bandura1986social}. Different events serve as distinct environmental stimuli, triggering varied cognitive patterns and behavioral responses. This triadic reciprocal causation mechanism suggests that depression expressions under different events comprise two information types: cross-event shared general information (public domain) reflecting core depression manifestations, and event-specific contextual information (private domain) capturing individualized responses to specific contexts. Effective disentanglement of these domains is crucial for reliable multi-event depression detection, as failure to distinguish them may result in distribution shift problems when encountering unseen individuals or events.

In response to these challenges, we propose P$^3$HF (Personality-guided Public-Private Domain Disentangled Hypergraph-Former Network), which integrates three key innovations: \\
\textbf{To address Problem 1}, a personality-guided feature regulation mechanism that enhances individual difference perception by utilizing LLM-generated personality descriptions to guide audiovisual feature extraction; \\
\textbf{To address Problem 2}, a novel Hypergraph-Former architecture that introduces positional encoding and attention mechanisms into hypergraph networks, effectively capturing both modal interactions and temporal dependencies; \\
\textbf{To address Problem 3}, an event-level domain disentanglement mechanism based on contrastive learning that distinguishes public and private domains across events, improving generalization capability and robustness in multi-event scenarios.

\section{Related Works}
\subsection{Individual-Aware in Depression Detection}
Modeling individual differences is a critical component in automated depression detection systems. Early approaches primarily relied on static demographic features as auxiliary information. \citet{kanchapogu2025deep} demonstrated effective bipolar and unipolar depression detection through joint modeling of structured demographic features with time-series behavioral data, while also introducing multi-task learning strategies that employ gender classification as an auxiliary task to enhance depression detection performance. Similarly, \citet{zhang2023deep} improved emotion recognition accuracy through gender-specific acoustic features, highlighting the importance of demographic considerations.
Recent advances have shifted toward incorporating psychological constructs, particularly personality traits. \citet{tan2025psy} proposed a psychology-informed module that validates the effectiveness of personality-aware representations in language tasks, achieving significant performance gains in depression detection scenarios. \citet{zhao2018personality} and \citet{fu2022adversarial} emphasized personality differences' profound impact on emotion recognition, proposing a personality-aware personalized framework. However, these methods exhibit fundamental limitations: discrete label representations of individual attributes fail to capture fine-grained individual feature differences, limiting their ability to model the heterogeneous nature of depression manifestations across different individuals.

\subsection{Graph-Based Multimodal Fusion}
Graph Neural Networks have demonstrated exceptional capability in modeling complex relational structures within multimodal data for mental health analysis. \citet{ghosal2019dialoguegcn} pioneered DialogueGCN, which models dialogues as speaker-centered graph structures to propagate emotional context between utterances. Building on this foundation, \citet{chen2022ms2} proposed MS$^{2}$-GNN for depression screening, jointly optimizing modal-shared and modal-specific graph branches to capture both common and unique cross-modal patterns.
Recent developments have explored hypergraph architectures for higher-order relationship modeling. \citet{yi2024multimodal} introduced the MFHACL model, combining hypergraph autoencoders with cross-modal contrastive learning to capture higher-order cross-modal interactions while explicitly aligning modalities in latent emotional representation space. DepressionMIGNN~\cite{zhao2025depressionmignn} combines RGCN and GAT approaches for utterance-level feature extraction across different modalities. While these hypergraph-based methods excel at modeling complex cross-modal relationships through unordered hyperedges, they inherently sacrifice temporal relationship modeling capabilities, which is a critical limitation for depression detection.DIB-HGCN~\cite{chen2025dynamic} constructed adaptive dialogue and monologue hyperedges to track cross-modal emotional changes in conversations. 

\subsection{Disentanglement and Representation Learning}
Domain feature disentanglement aims to separate domain-invariant and domain-specific information in learned representations. \citet{bengio2013representation} first formalized disentangled representation learning, arguing that multiple explanatory factors are often mixed within representations. \citet{locatello2019challenging} advanced this field through explicit modeling of data generation processes, while \citet{zellinger2019robust} proposed robust unsupervised methods for domain-invariant representation learning through distribution alignment.
Graph-based and multimodal approaches have embraced disentanglement principles. AM-GCN~\cite{wang2020gcn} extracts specific and shared embeddings from node features and topological structures, while MISA~\cite{hazarika2020misa} learns modal-invariant and modal-specific representations for sentiment analysis. \citet{sun2023modality} proposed gated cross-modal attention mechanisms for filtering cross-modal inconsistencies, and \citet{ravi2024enhancing} introduced speaker disentanglement in depression detection to remove speaker-specific characteristics.
However, existing approaches primarily address multi-modal or individual difference disentanglement, with limited consideration for event-level disentanglement. In mental health detection, individuals exhibit distinct behavioral patterns across different events, necessitating explicit modeling of event-specific versus event-invariant depression manifestations.

\section{Methodology}

\begin{figure*}[t]
\centering
\includegraphics[width=1\textwidth]{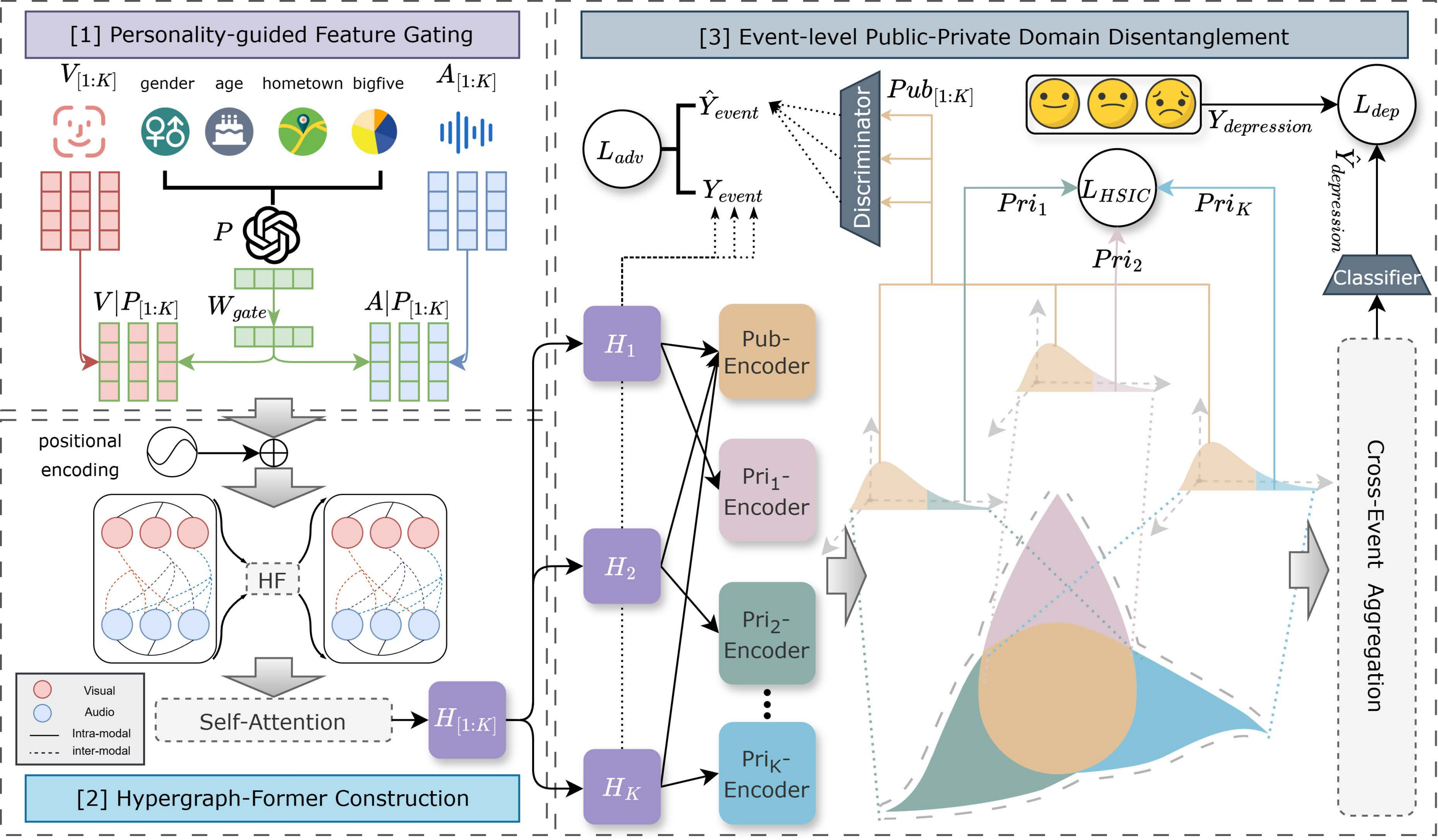}
\caption{Model architecture of our P$^3$HF, which is optimized through a combinational objective of three parts. Since the number of events in the dataset we use is three, the legend is drawn in the form of $K=3$.}
\label{fig:model}
\end{figure*}

\paragraph{Overview.}
Given sample $S$, composed of $K$ different events, $\{E_1, E_2, \ldots, E_K\}$. $E_k$ has different lengths $T_k$, indicating the number of segmented frames. $E_k = \{V_k, A_k\}$,  representing visual and audio modalities respectively. For $V_k$ and $A_k$, $V_k = \{v_1, v_2, \ldots, v_{T_k}\}$, $A_k = \{a_1, a_2, \ldots, a_{T_k}\}$, indicating that two modalities in the same sample across different events are divided into $T_k$ segments. To align different event lengths, we adopt a repetition-based padding strategy. Each sample $S$ includes a unique depression label $Y_{\text{depression}}$, Big Five personality(neuroticism, extraversion, openness, agreeableness, and conscientiousness) scores, and demographic information (gender, age, and hometown), which are known to be associated with individual personality traits. In automated depression detection tasks, our objective is to construct a function $\mathcal{F}$ predicting depression labels for each segment. As mentioned in the \textit{Introduction}, to address the mentioned problems, we first embed individual information $P$ as prior into audiovisual features. Then, modalities from event $k$ undergo fusion $\mathcal{G}$ to obtain $H_k$, and contrastive learning is applied to obtain the final prediction results: 
\begin{align}
\hat{y} = \mathcal{F}(&\mathcal{G}(V_1 | P, A_1 | P),\ \mathcal{G}(V_2 | P, A_2 | P), \notag \\
               &\ldots,\ \mathcal{G}(V_K | P, A_K | P)), k \in [1, K]
\end{align}

\paragraph{Preprocessing.}
For each sample's visual feature extraction, we employ Haar Cascade face detection, then use pretrained ResNet-50 \cite{he2016deep} to extract 2048-dimensional features $V_k \in \mathbb{R}^{T_k \times 2048}$. Audio features $A_k \in \mathbb{R}^{T_k \times 1024}$ are extracted using Chinese fine-tuned Wav2Vec2 \cite{baevski2020wav2vec} model. For personality information $P$, we construct prompts containing gender, age, hometown, and Big Five personality scores as GPT-4 input (temperature=0 for reproducibility), obtaining descriptive text capturing individual characteristics. We then use BERT \cite{devlin2019bert} to encode generated personality descriptions, producing 768-dimensional text features.

\paragraph{Personality-guided Feature Gating.}
As shown in the first part of Figure~\ref{fig:model}, our method adopts multimodal inputs, including visual features $V_{[1:K]}$, audio features $A_{[1:K]}$, and personality information $P$ preprocessed from the MPDD dataset. To capture contextual information, we apply bidirectional LSTM layers to $V_k$, $A_k$, and $P$, unifying all feature dimensions to $D_1$:
\begin{align}
\tilde{V}_k &= \text{Bi-LSTM}(V_k) \in \mathbb{R}^{T_k \times D_1} \label{eq:bv}\\
\tilde{A}_k &= \text{Bi-LSTM}(A_k) \in \mathbb{R}^{T_k \times D_1} \label{eq:ba}\\
\tilde{P} &= \text{Bi-LSTM}(P) \in \mathbb{R}^{D_1} \label{eq:bp}
\end{align}
To derive adaptive gating weights conditioned on different individuals, we apply a learnable linear transformation to the personalized representation $\tilde{P}$:
\begin{align}
W_{\text{gate}} = \sigma\left( \mathbf{W}_p \tilde{P} + \mathbf{b}_p \right) \in \mathbb{R}^{D_1}
\label{eq:wg}
\end{align}
where $\sigma(\cdot)$ denotes the sigmoid activation function, $\mathbf{W}_p$ and $\mathbf{b}_p$ are learnable parameters.

Inspired by ResNet \cite{he2016deep}, we introduce residual connections to prevent gradient vanishing and regulate audio and visual features through gating mechanism:
\begin{align}
A_k|P &= \tilde{A}_k + \tilde{A}_k \odot W_{\text{gate}}, & 
V_k|P &= \tilde{V}_k + \tilde{V}_k \odot W_{\text{gate}}
\label{eq:gate}
\end{align}
where $\odot$ represents the multiplication of elements in a broadcasting mechanism, $W_{gate} \in \mathbb{R}^{D_1}$ is broadcast to match the temporal dimension $T_k$. This produces personality-guided audiovisual features $A|P$ and $V|P$, representing features modulated by prior individual information.

\paragraph{Hypergraph-Former Construction.}
To address the lack of temporal relationships in traditional hypergraphs, we integrate sinusoidal positional encoding into personality-guided features. For each event $k$, we add positional encoding to audio and visual features:
\begin{align}
\hat{A}_k &= A_k|P + \text{PE}(A_k|P) \in \mathbb{R}^{T_k \times D_1} \label{eq:hfb}\\
\hat{V}_k &= V_k|P + \text{PE}(V_k|P) \in \mathbb{R}^{T_k \times D_1}
\end{align}
where $PE(\cdot)$ represents sinusoidal positional encoding, injecting temporal order information into feature representations.

We then construct hypergraph $\mathcal{H} = (\mathcal{V}, \mathcal{E})$ for each event, where node set $\mathcal{V}$ contains all audio and visual features from the same sample and event, totaling $2T_k$ nodes. Hyperedge set $\mathcal{E}$ is constructed using predefined sliding windows of size $w$.  The windowed construction strategy is motivated by observations that depression-related patterns frequently exhibit local temporal consistency, where adjacent time steps share contextual information crucial for accurate detection. To capture high-order intra-modal and inter-modal relationships within each window, we create hyperedges by: (1) connecting all nodes of the same modality within windows to enhance intra-modal local features (represented by solid lines in part two of Figure~\ref{fig:model}); (2) connecting each node of one modality to all nodes of another modality within windows to model local inter-modal interactions (represented by dashed lines in part two of Figure~\ref{fig:model}). This produces $(T_k - w + 1) \times (2 + 2w)$ hyperedges total, comprehensively covering temporal and inter-modal relationships.
Following the hypergraph neural network framework, we compute node representations through hypergraph convolution. Hypergraph convolution operations aggregate information from connected nodes through hyperedges:
\begin{align}
X^{(l+1)} &= \sigma(\mathbf{D}_v^{-1/2} \mathbf{H} \mathbf{W}_e \mathbf{D}_e^{-1} \mathbf{H}^T \mathbf{D}_v^{-1/2} X^{(l)} \mathbf{\Theta}^{(l)})
\end{align}
where $\mathbf{H} \in \mathbb{R}^{|\mathcal{V}| \times |\mathcal{E}|}$ is the incidence matrix, $\mathbf{D}_v$ and $\mathbf{D}_e$ are diagonal degree matrices for nodes and hyperedges respectively, $\mathbf{W}_e$ denotes the hyperedge weight matrix, $\mathbf{\Theta}^{(l)}$ is the learnable parameter matrix. 
Output dimensions are set to $D_2$ for improved computational efficiency.

To enhance feature interactions and capture global dependencies beyond local hypergraph connections, we apply multi-head self-attention ($M$) to hypergraph-processed features:
\begin{align}
M = \bigoplus_{i=1}^h \left\{\mathrm{softmax}\left( 
\frac{\mathbf{Q}\mathbf{W^Q_i}(\mathbf{K}\mathbf{W^K_i})^T}{\sqrt{D_2}} 
\right) \mathbf{V}\mathbf{W^V_i} \right\}\mathbf{W^O}
\end{align}
where $\mathbf{Q}, \mathbf{K}, \mathbf{V}$ represent the audio or visual features processed by the hypergraph module, $\mathbf{W^Q_i}$, $\mathbf{W^K_i}$, $\mathbf{W^V_i}$, $\mathbf{W^O}$ are learnable projection matrices, $h$ is the number of attention heads. This operation is applied separately to both audio and visual features to obtain $\tilde{A}_k, \tilde{V}_k \in \mathbb{R}^{T_k \times D_2}$.

Finally, we concatenate the attention-enhanced features to obtain unified representations:

\begin{align}
H_k = \bm{\oplus}(A_k^{(\text{att})}, V_k^{(\text{att})}) \in \mathbb{R}^{T_k \times 2D_2} \label{eq:hfe}
\end{align}

\paragraph{Public-Private Domain Disentanglement.}
As shown in part three of Figure~\ref{fig:model}, Hypergraph-Former outputs $[H_1, H_2, \dots, H_K]$ represent features from $K$ events respectively. Inspired by \citet{zellinger2019robust}, we aim to learn domain-invariant representations across different events while learning individual event-specific features. To capture public distributions across events, we input all hypergraph representations into a shared public encoder:
\begin{align}
Pub_k = \text{Pub-Enc}(H_k) \in \mathbb{R}^{T_k \times D_3}
\label{eq:pub}
\end{align}
Meanwhile, since we have previously obtained Individual guided features, to model private distributions across different individuals and different events, we use independent private encoders for each event:
\begin{align}
Pri_k = \text{Pri}_k\text{-Enc}(H_k) \in \mathbb{R}^{T_k \times D_3}
\label{eq:pri}
\end{align}

Subsequently, based on contrastive learning paradigms, we pull public features together and push private features apart. 
For the public domain, we adopt adversarial training following the GAN framework, where our public encoder acts as the generator producing $Pub_k$, while a discriminator is trained to predict event labels from these representations:
\begin{align}
\hat{Y}_{\text{event},k} = \text{Disc}(Pub_k) \in \mathbb{R}^{T_k}
\end{align}
The discriminator loss encourages accurate event classification, continuously improving the discriminator's performance through cross-entropy optimization:
\begin{align}
\mathcal{L}_{\text{disc}} &= -\sum_{k=1}^{K}\sum_{t=1}^{Tk} Y_{\text{event},k,t} \log \hat{Y}_{\text{event},k,t}
\label{eq:ld}
\end{align}
where $Y_{\text{event}}$ represents the actual event sequence numbers of the true sources of various features, encoded as one-hot vectors.
To effectively disentangle public features that are invariant across events, we formulate the adversarial training as a MinMax optimization problem:
\begin{align}
\mathcal{L}_{\text{adv}} &= \min_{\text{Pub-Enc}} \max_{\text{Disc}} \sum_{k=1}^{K}\sum_{t=1}^{Tk} Y_{\text{event},k,t} \log \text{Disc}(\text{Pub-Enc}(H_k))_t 
\label{eq:la}
\end{align}

For the private domain, we adopt Hilbert-Schmidt Independence Criterion (HSIC) to measure independence between private representations from different events. HSIC quantifies dependence between two random variables by computing the Hilbert-Schmidt norm of their cross-covariance operator. Given private representations $Pri_i$ and $Pri_j$, HSIC is computed as, with HSIC approaching 0 indicating greater variable independence:
\begin{align}
\text{HSIC}(Pri_i, Pri_j) = \text{trace}(L_i H L_j H)
\end{align}
where $L_i$ and $L_j$ are kernel matrices computed using RBF kernels ($\sigma=1.0$), and $H$ is the centering matrix. By minimizing $\mathcal{L}_{HSIC} = \sum_{i \neq j} \text{HSIC}(Pri_i, Pri_j)$, we ensure private encoders capture event-specific features while maintaining independence across different events.
Finally, we concatenate averaged public representations with all private representations to aggregate multi-event features:
\begin{align}
I = \bigoplus\left\{\frac{1}{K}\sum_{k=1}^{K}Pub_k, Pri_1, \dots, Pri_K\right\} \in \mathbb{R}^{T_k \times (1+K)D_3}
\label{eq:i}
\end{align}

\paragraph{Depression Detection.}
We project $I$ through multiple linear layers to obtain the final output $\hat{Y}_{\text{depression}} \in \mathbb{R}^{T_k \times 3}$, where the three classes represent normal, mildly depressed, and severely depressed states respectively.
The depression classification loss employs negative log-likelihood (NLL) loss:
\begin{align}
\mathcal{L}_{\text{dep}} = -\frac{1}{T_k}\sum_{t=1}^{T_k} \log P(\hat{Y}_{\text{depression},t} = Y_{\text{depression},t})
\label{eq:lt}
\end{align}
The overall training objective combines all loss components:

\begin{align}
\mathcal{L}_{\text{main}} = \alpha \mathcal{L}_{\text{dep}} + \beta \mathcal{L}_{\text{adv}} + \gamma \mathcal{L}_{\text{HSIC}}
\end{align}
with weight constraints $\alpha + \beta + \gamma = 1$ ensuring balanced optimization.
We employ an alternating training strategy where the discriminator aims to minimize $\mathcal{L}_{\text{disc}}$ (maximize $\mathcal{L}_{\text{adv}}$) while the main model minimizes $\mathcal{L}_{\text{main}}$, as detailed in Algorithm~\ref{alg:p3hf_training}. The optimal discriminator accuracy around $1/3$ indicates successful public domain learning.

{\tiny
\begin{algorithm}[h]
\caption{Training Process of P$^3$HF}
\label{alg:p3hf_training}
\begin{algorithmic}[1]
\REQUIRE Dataset with $K$ events $\{E_1, \ldots, E_K\}$, personality $P$, labels $Y_{\text{depression}}$
\REQUIRE Hyperparameters $D_1, D_2, D_3, \alpha, \beta, \gamma$
\ENSURE Trained P$^3$HF model for depression detection

\STATE Initialize network components: encoders, Hypergraph-Former parameters, discriminator, classifier
\FOR{each training epoch}
    \FOR{each mini-batch}
        \STATE \textbf{// Forward Pass}
        \STATE $\tilde{U} \leftarrow  \text{Bi-LSTM}(U), M_k \in \{V_k, A_k, P\}$ \textbf{// Eq.\ref{eq:bv}-\ref{eq:bp}} 
        \STATE $W_{\text{gate}} \leftarrow \sigma(W_p \tilde{P} + b_p)$
        \FOR{$k = 1$ to $K$}
            \STATE $A_k|P, V_k|P \leftarrow \text{Gating}(A_k, V_k, W_{\text{gate}})$ \textbf{// Eq.\ref{eq:gate}}
            \STATE $H_k \leftarrow \text{HF}(A_k|P, V_k|P)$ \textbf{// Eq.\ref{eq:hfb}-\ref{eq:hfe}}
            \STATE $Pub_k, Pri_k\leftarrow \text{Domain}(H_k)$ \textbf{// Eq.\ref{eq:pub}-\ref{eq:pri}}
        \ENDFOR
        \STATE $I \leftarrow \bm{\oplus}(\frac{\sum_{k=1}^{K}Pub_k}{K}, Pri_1, \ldots, Pri_K)$ \textbf{// Eq.\ref{eq:i}}
        \STATE $\hat{Y}_{\text{event}} \leftarrow \text{Disc}(\{Pub_{[1:K]}\})$
        \STATE $\hat{Y}_{\text{depression}} \leftarrow \text{Classifier}(I)$
        \STATE \textbf{// Loss Computation}
        \STATE $\mathcal{L}_{\text{disc}} \leftarrow \text{CE}(Y_{\text{event}}, \hat{Y}_{\text{event}})$; $\mathcal{L}_{\text{adv}} \leftarrow -\mathcal{L}_{\text{disc}}$ \textbf{// Eq.\ref{eq:ld}-\ref{eq:la}}
        \STATE $\mathcal{L}_{\text{HSIC}} \leftarrow \sum_{i \neq j} \text{HSIC}(Pri_i, Pri_j)$ 
        \STATE $\mathcal{L}_{\text{dep}} \leftarrow \text{NLL}(Y_{\text{depression}}, \hat{Y}_{\text{depression}})$ \textbf{// Eq.\ref{eq:lt}}
        \STATE \textbf{// Alternating Adversarial Training}
        \STATE \textbf{Step 1:} $\theta_{\text{disc}} \leftarrow \theta_{\text{disc}} - \eta_{\text{disc}} \nabla \mathcal{L}_{\text{disc}}$
        \STATE \textbf{Step 2:} $\theta_{\text{main}} \leftarrow \theta_{\text{main}} - \eta_{\text{main}} \nabla \mathcal{L}_{\text{main}}$
        
    \ENDFOR
\ENDFOR
\end{algorithmic}
\end{algorithm}
}

\section{Experiment}
\subsection{Experiment Protocol}
\paragraph{Dataset.}
To investigate the effectiveness of personality-guided multimodal multi-event models, we conduct experiments on the MPDD-Young dataset from the MPDD Challenge. This dataset contains multimodal recordings (audio and video) from young populations, annotated with PHQ-9 depression scores and multi-dimensional individual information including age, gender, hometown, and Big Five personality traits. Data collection involves subjects performing three tasks (i.e., $K=3$): self-introduction and two text reading tasks, capturing subjects' multimodal manifestations under natural and guided contexts. 
\textbf{It should be noted that among existing publicly available depression detection datasets, MPDD is currently the only dataset simultaneously possessing multi-event structure, multimodal recording, and multi-dimensional individual information labels.} This makes it uniquely suitable for our proposed architecture.

\paragraph{Evaluation Metrics.}
We evaluate model performance using two commonly used performance metrics for each subject's final prediction results in both binary and three-class classification tasks: weighted F1-score (w-F1) and accuracy (Acc). Accuracy measures the overall proportion of correct model predictions; weighted F1-score comprehensively considers precision and recall for each class, weighted averaged by the number of samples in each class, making it more suitable for class imbalance scenarios.

\paragraph{Implementation Details.}

All experiments in this study are conducted on a Windows 10 system with an NVIDIA RTX 4090 GPU, implemented using PyTorch 1.13.1 and PyG 2.6.1 frameworks. We set batch size to 20, with maximum training of 300 epochs per experiment. Learning rate adopts cosine annealing strategy (1e-4 to 1e-5), optimized via Optuna. The optimizer uses Adam with weight decay of 5e-4. To prevent overfitting, we introduce early stopping and checkpoint mechanisms to save optimal models. Experiments were repeated with 10 random seeds, showing significant differences (one-way ANOVA, p $<$ 0.05) across method variants.

\subsection{Performance Comparison}

We evaluate P$^3$HF against state-of-the-art depression detection methods across unimodal and multimodal paradigms on MPDD-Young dataset. Baselines include: (i) \textbf{Unimodal}: NUSD~\cite{wang2023non} employs non-uniform processing for speaker-invariant speech analysis; STA-DRN~\cite{pan2024spatial} leverages spatial-temporal attention for facial expression dynamics. (ii) \textbf{Multimodal}: Gated LSTM~\cite{rohanian2019detecting} for word-level fusion; TBN~\cite{kazakos2019epic} adapts EPIC-Fusion for audiovisual temporal modeling; IA fusion~\cite{chumachenko2022self} handles incomplete multimodal data via self-attention; DEP-Former~\cite{ye2024dep} analyzes emotional changes through Transformer architecture; MGLRA~\cite{meng2024masked} combines masked graph learning with recurrent alignment; DepMamba~\cite{ye2025depmamba} utilizes progressive Mamba-based fusion; MPDD baseline~\cite{fu2025first} integrates personalized features.

\begin{table}[t]
\centering\small
\setlength{\tabcolsep}{8pt}
\begin{tabular}{l||cc|cc}
\toprule
\multirow{2}{*}{\textbf{Method}} & \multicolumn{2}{c|}{\textbf{Binary}} & \multicolumn{2}{c}{\textbf{Ternary}} \\
\cmidrule(lr){2-5} & ACC & w-F1 & ACC & w-F1 \\
\midrule
\rowcolor{gray!20}
\multicolumn{5}{l}{\small\textit{Unimodal}} \\
NUSD (2023) & 63.01 & 60.64 & 57.19 & 55.44 \\
STA-DRN (2024) & 64.14 & 62.23 & 58.93 & 57.34 \\
\rowcolor{gray!20}
\multicolumn{5}{l}{\small\textit{Multimodal}} \\
Baseline (2025) & 63.64 & 59.96 & 49.66 & 51.86 \\
Gated LSTM (2019) & 64.48 & 62.17 & 52.51 & 50.32 \\
TBN (2019) & 66.21 & 64.77 & 61.76 & 60.23 \\
IA fusion (2022) & 68.41 & 67.23 & 62.87 & 61.39 \\
DEP-Former (2024) & 67.85 & 66.23 & 63.43 & 61.75 \\
MGLRA (2024) & 70.37 & 68.93 & 61.35 & 59.78 \\
DepMamba (2025) & 72.56 & 71.44 & 67.85 & 66.23 \\
\midrule
\textbf{P$^3$HF (Ours)} & \textbf{82.17} & \textbf{81.39} & \textbf{76.29} & \textbf{74.61} \\
\bottomrule
\end{tabular}
\caption{Comparative performance on MPDD-Young dataset. Our method achieves substantial improvements across both classification tasks.}
\label{tab:comparison}
\end{table}

\paragraph{Quantitative Analysis.} Table~\ref{tab:comparison} demonstrates P$^3$HF's superior performance. For binary classification, we achieve 82.17\%/81.39\% (ACC/w-F1), surpassing the strongest baseline DepMamba by 9.61\%/9.95\%. Ternary classification shows consistent improvements of 8.44\%/8.38\% over DepMamba (76.29\%/74.61\% vs. 67.85\%/66.23\%), validating our approach's effectiveness for fine-grained depression severity assessment.

\paragraph{Architectural Advantages.} Our superior performance stems from three key innovations: (i) \textbf{Individual Modeling}: LLM-based personality embeddings enable personalized depression pattern adaptation, addressing individual symptom expression variability ignored by traditional approaches (Gated LSTM, IA fusion, STA-DRN). (ii) \textbf{Multimodal Integration}: Hypergraph-Former captures high-order intra/cross-modal relationships with temporal awareness, overcoming limitations of existing methods—MGLRA lacks temporal modeling while TBN cannot handle high-order dependencies. (iii) \textbf{Multi-event Generalization}: Domain disentanglement mechanisms distinguish context-specific manifestations, addressing critical gaps in current methods—NUSD focuses solely on speech disentanglement, DepMamba's progressive fusion struggles with cross-event feature discrimination, and DEP-Former's emotional analysis lacks multi-event contextual understanding.

The substantial performance gaps (8.38\%-9.95\% improvements) demonstrate that personality-guided, hypergraph-enhanced multimodal fusion with domain disentanglement addresses fundamental limitations in existing depression detection paradigms, particularly for cross-event generalization and individual adaptation.

\begin{table}[t]
\centering\small
\begin{tabular}{l||cc|cc}
\toprule
\multirow{2}{*}{\textbf{Component}} & \multicolumn{2}{c|}{\textbf{Binary}} & \multicolumn{2}{c}{\textbf{Ternary}} \\
\cmidrule(lr){2-5} & ACC & w-F1 & ACC & w-F1 \\
\midrule
\rowcolor{gray!20}
\multicolumn{5}{l}{\small\textit{Multimodal Fusion}} \\
w/o visual & 77.52 & 76.63 & 72.94 & 70.52 \\
w/o audio & 76.89 & 75.77 & 70.85 & 69.39 \\
\rowcolor{gray!20}
\multicolumn{5}{l}{\small\textit{Domain Disentanglement}} \\
w/o domain disentanglemen & 71.84 & 70.17 & 66.53 & 65.72 \\
w/o pub-domain & 75.34 & 74.38 & 70.30 & 68.19 \\
w/o pri-domain & 78.15 & 77.02 & 74.01 & 71.32 \\
\rowcolor{gray!20}
\multicolumn{5}{l}{\small\textit{Personality Guidance}} \\
w/o personal information & 76.68 & 75.41 & 71.55 & 69.24 \\
w/ numeric embedding & 80.61 & 78.77 & 75.32 & 73.34 \\
\midrule
\textbf{Full Model} & \textbf{82.17} & \textbf{81.39} & \textbf{76.29} & \textbf{74.61} \\
\bottomrule
\end{tabular}
\caption{Component ablation results demonstrating each module's contribution.}
\label{tab:component}
\end{table}

\begin{table}[t]
\centering\small
\begin{tabular}{l||cc|cc}
\toprule
\multirow{2}{*}{\textbf{Architecture}} & \multicolumn{2}{c|}{\textbf{Binary}} & \multicolumn{2}{c}{\textbf{Ternary}} \\
\cmidrule(lr){2-5} & ACC & w-F1 & ACC & w-F1 \\
\midrule
Cross-Attention & 75.82 & 74.59 & 69.15 & 67.33 \\
Directed GCN & 77.55 & 76.24 & 72.41 & 69.29 \\
Undirected GCN & 79.33 & 78.17 & 73.57 & 71.94 \\
Directed GAT & 78.51 & 77.42 & 72.82 & 70.68 \\
Undirected GAT & 80.07 & 79.14 & 74.23 & 72.51 \\
Hypergraph & 79.68 & 78.73 & 73.86 & 72.05 \\
Hypergraph-Attention & 80.31 & 79.55 & 74.32 & 72.94 \\
\midrule
\textbf{Hypergraph-Former} & \textbf{82.17} & \textbf{81.39} & \textbf{76.29} & \textbf{74.61} \\
\bottomrule
\end{tabular}
\caption{Architectural comparison revealing Hypergraph-Former's superiority.}
\label{tab:HF}
\end{table}

\subsection{Ablation Study}
\paragraph{Component-wise Analysis.}
We conduct comprehensive ablation experiments to quantify each component's contribution in P$^3$HF. Table~\ref{tab:component} presents results on MPDD dataset, revealing critical insights into architectural design choices.
Removing visual modality causes 4.65\% accuracy drop (binary) and 3.35\% drop (ternary), while audio removal leads to 5.28\% and 5.44\% degradation respectively. The asymmetric impact suggests audio features capture more discriminative temporal patterns for personalized correlations, aligning with psychological theories emphasizing prosodic cues in mental health assessment.
The public domain encoder removal severely impacts performance (6.83\%/5.99\%), demonstrating its crucial role in extracting event-invariant representations. Private domain removal shows smaller degradation (4.02\%/2.28\%), indicating event-specific features provide complementary but less critical information. This asymmetry validates our hypothesis that shared personality traits dominate individual patterns.
Personal information removal causes consistent degradation (5.49\%/4.74\%), confirming that personality-guided attention effectively captures individual differences. The complete model achieves 82.17\%/81.39\% (binary) and 76.29\%/74.61\% (ternary) performance, substantially outperforming all ablated versions.

\paragraph{Architectural Comparison.}
Table~\ref{tab:HF} compares alternative architectures against Hypergraph-Former. Cross-attention exhibits limited capability (75.82\%/69.15\%), failing to model complex multimodal dependencies. GNNs show progressive improvement: directed GCN (77.55\%/72.41\%) $<$ undirected GCN (79.33\%/73.57\%) $<$ undirected GAT (80.07\%/74.23\%), confirming bidirectional propagation and attention mechanisms' benefits.
Standard hypergraph achieves competitive results (79.68\%/73.86\%), but Hypergraph-Former surpasses it by 2.49\%/2.43\%, validating our key innovations: (i) positional encoding captures temporal dependencies, (ii) self-attention enhances local feature discrimination, and (iii) hypergraph structure models high-order cross-modal relationships.

\begin{figure}[t]
\centering
\includegraphics[width=1\columnwidth]{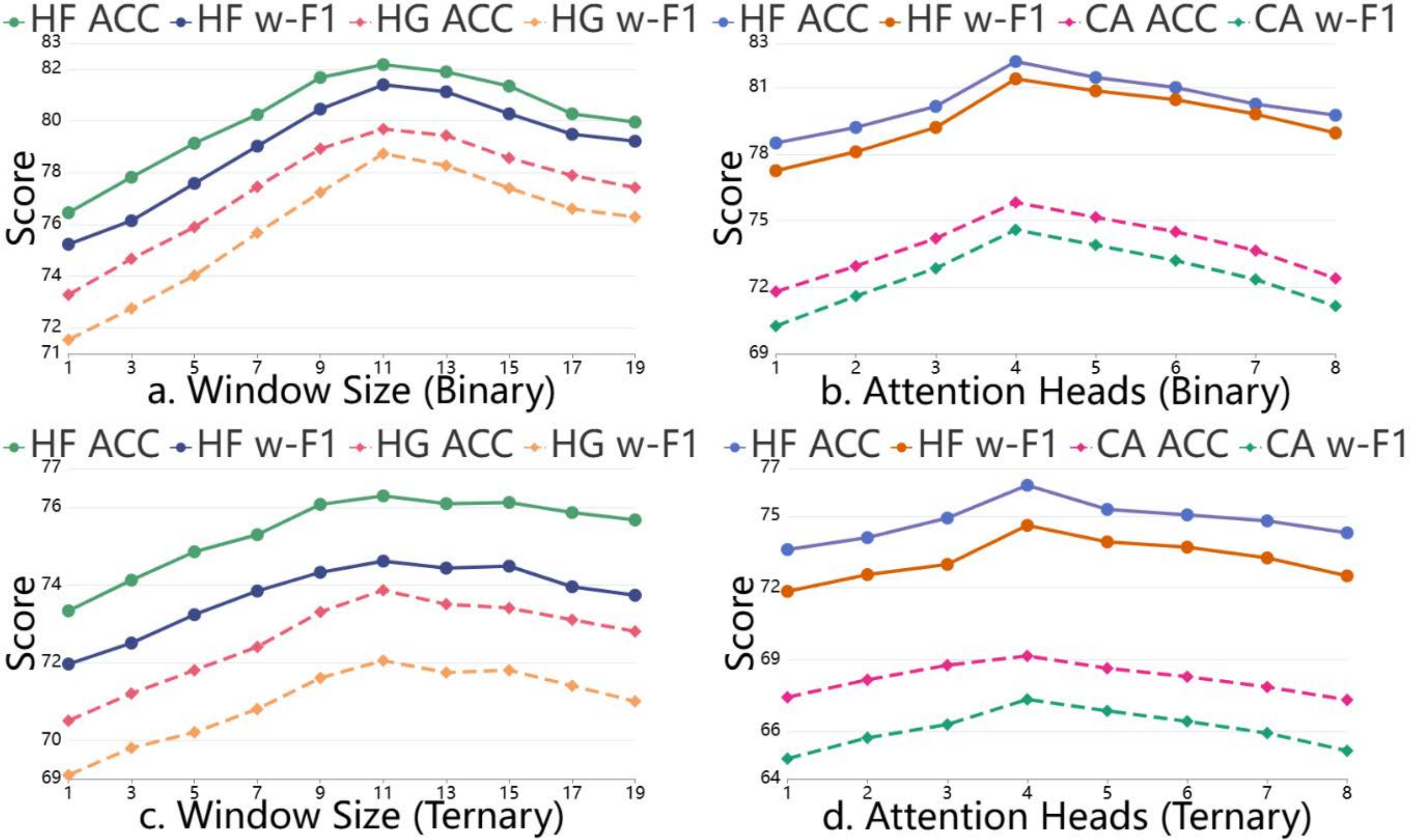}
\caption{Hyperparameter sensitivity analysis showing optimal configurations for window size and attention heads across binary/ternary tasks. HF:hypergraph-former; HG:hypergraph; CA:cross attention.}
\label{fig:line}
\end{figure}

\paragraph{Hyperparameter Sensitivity.}
Figure~\ref{fig:line} reveals critical hyperparameter dependencies. Window size exhibits inverted-U relationship: size=1 reduces hypergraph to simple connections, limiting local consistency modeling; optimal performance at size=11 balances temporal context and computational efficiency; larger windows introduce noise, degrading performance. This finding suggests individual patterns require moderate temporal context ($\approx$ 11 time steps) for optimal characterization.

Attention heads show similar patterns with optimum at 4 heads. Insufficient heads ($\leq 3$) limit cross-modal relationship modeling, while excessive heads ($\geq 5$) cause attention redundancy and overfitting. This reveals the inherent complexity of personalized  multimodal interactions requires precisely balanced attention diversity.

\begin{figure}[t]
\centering
\includegraphics[width=1\columnwidth]{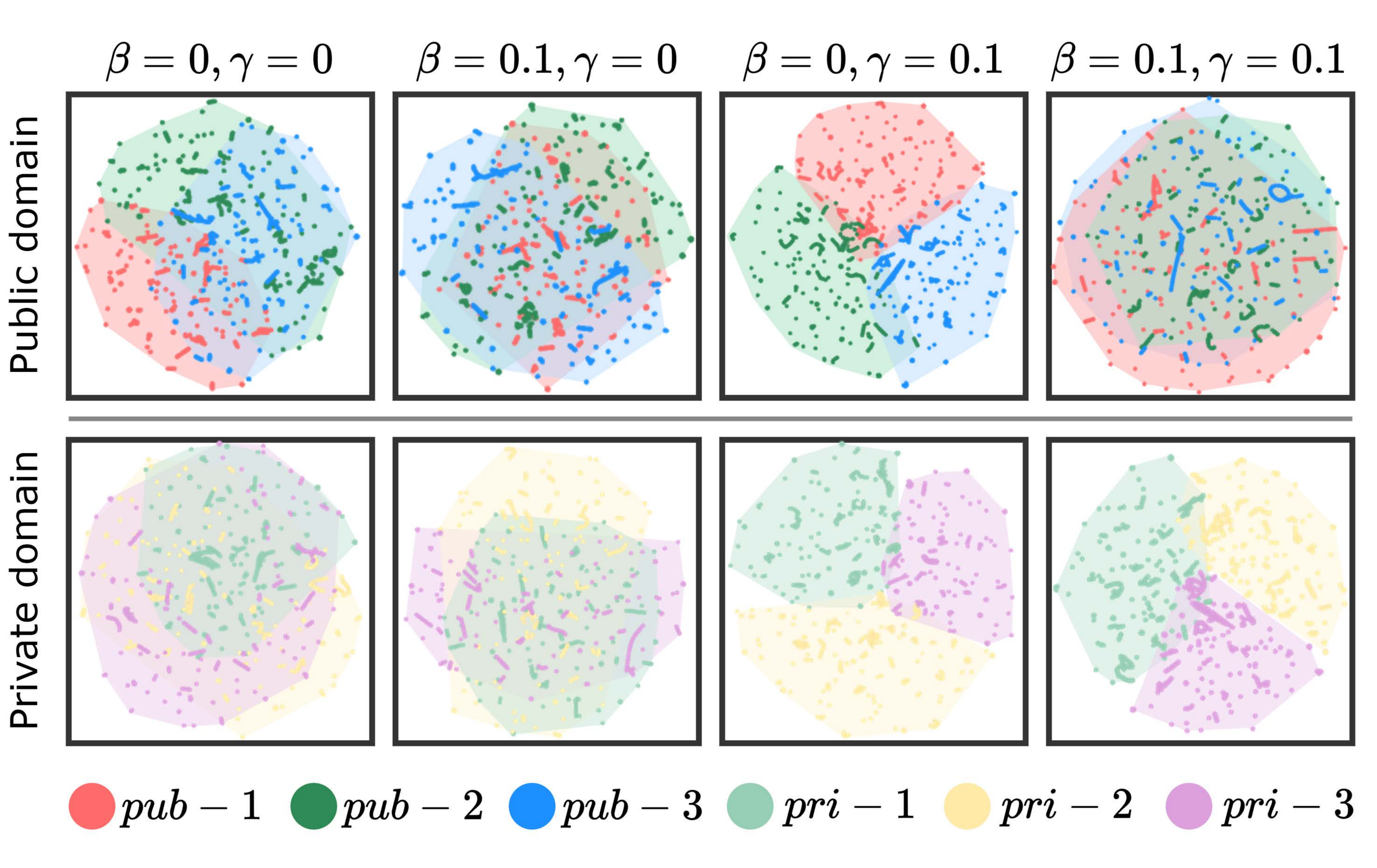}
\caption{Domain disentanglement visualization under different loss weights. Optimal configuration ($\beta=\gamma=0.1$) achieves clear public-private feature separation.}
\label{fig:vis}
\end{figure}

\paragraph{Domain Disentanglement Visualization.}
Figure~\ref{fig:vis} provides t-SNE visualization of domain disentanglement effects under varying loss configurations ($\beta$: adversarial loss, $\gamma$: HSIC loss). Without disentanglement ($\beta=\gamma=0$), features exhibit chaotic mixing across events. Partial optimization ($\beta=0.1,\gamma=0$ or $\beta=0,\gamma=0.1$) achieves incomplete separation. Optimal configuration ($\beta=\gamma=0.1$) demonstrates clear disentanglement: public features converge to unified distributions (event-invariant), while private features occupy distinct spaces (event-specific). This validates our theoretical framework that personality traits manifest as stable cross-event patterns while contextual factors remain event-dependent.

\section{Conclusion}
This paper proposes the P$^3$HF framework through personality-guided feature regulation, temporal-aware hypergraph modeling, and event-level domain disentanglement. On the MPDD-Young dataset, P$^3$HF achieves around 10\% improvement on accuracy  and weighted F1 for binary and ternary classification task, attaining state-of-the-art performance. Experimental results fully validate the effectiveness of our proposed innovations. We first introduce LLMs for multi-dimensional individual information description generation, breaking through traditional discrete label limitations and significantly improving model perception capabilities for different individuals; innovatively introduce positional encoding and attention mechanisms into hypergraph networks, effectively addressing traditional hypergraph deficiencies in temporal information modeling; first introduce event-level domain disentanglement mechanisms in depression detection, successfully modeling public-private domain distributions and effectively addressing distribution shift problems across multi-event scenarios. Additionally, our method is currently validated on depression detection tasks and holds promise for providing important methodological guidance for other mental health detection including anxiety disorders and bipolar affective disorders, while exploring more general model applications in related directions.

\section{Acknowledgments}
This research is supported by the National Natural Science Foundation of China (Grant No. 62306068) Project, the Natural Science Foundation of Hebei Province, China (Grant No. F2024501002), and the Fundamental Research Funds for the Central Universities (Grant No. N2523005).


\begin{thebibliography}{51}
\providecommand{\natexlab}[1]{#1}

\bibitem[{Baevski et~al.(2020)Baevski, Zhou, Mohamed, and Auli}]{baevski2020wav2vec}
Baevski, A.; Zhou, Y.; Mohamed, A.; and Auli, M. 2020.
\newblock wav2vec 2.0: A framework for self-supervised learning of speech representations.
\newblock \emph{Advances in neural information processing systems}, 33: 12449--12460.

\bibitem[{Bandura(1986)}]{bandura1986social}
Bandura, A. 1986.
\newblock \emph{Social Foundations of Thought and Action: A Social Cognitive Theory}.
\newblock Prentice-Hall.
\newblock ISBN 978-0-13-815614-5.

\bibitem[{Bengio, Courville, and Vincent(2013)}]{bengio2013representation}
Bengio, Y.; Courville, A.; and Vincent, P. 2013.
\newblock Representation learning: A review and new perspectives.
\newblock \emph{IEEE transactions on pattern analysis and machine intelligence}, 35(8): 1798--1828.

\bibitem[{Chen et~al.(2022)Chen, Hong, Guo, Hao, and Hu}]{chen2022ms2}
Chen, T.; Hong, R.; Guo, Y.; Hao, S.; and Hu, B. 2022.
\newblock MS$^2$-GNN: Exploring GNN-based multimodal fusion network for depression detection.
\newblock \emph{IEEE Transactions on Cybernetics}, 53(12): 7749--7759.

\bibitem[{Chen and Shi(2025)}]{chen2025dynamic}
Chen, X.; and Shi, W. 2025.
\newblock Dynamic Interactive Bimodal Hypergraph Networks for Emotion Recognition in Conversations.
\newblock In \emph{Proceedings of the AAAI Conference on Artificial Intelligence}, volume~39, 1256--1264.

\bibitem[{Chen et~al.(2024)Chen, Deng, Zhou, Wu, Qian, and Huang}]{chen2024depression}
Chen, Z.; Deng, J.; Zhou, J.; Wu, J.; Qian, T.; and Huang, M. 2024.
\newblock Depression detection in clinical interviews with LLM-empowered structural element graph.
\newblock In \emph{Proceedings of the 2024 Conference of the North American Chapter of the Association for Computational Linguistics: Human Language Technologies (Volume 1: Long Papers)}, 8181--8194.

\bibitem[{Chumachenko, Iosifidis, and Gabbouj(2022)}]{chumachenko2022self}
Chumachenko, K.; Iosifidis, A.; and Gabbouj, M. 2022.
\newblock Self-attention fusion for audiovisual emotion recognition with incomplete data.
\newblock In \emph{2022 26th International Conference on Pattern Recognition (ICPR)}, 2822--2828. IEEE.

\bibitem[{Devlin et~al.(2019)Devlin, Chang, Lee, and Toutanova}]{devlin2019bert}
Devlin, J.; Chang, M.-W.; Lee, K.; and Toutanova, K. 2019.
\newblock Bert: Pre-training of deep bidirectional transformers for language understanding.
\newblock In \emph{Proceedings of the 2019 conference of the North American chapter of the association for computational linguistics: human language technologies, volume 1 (long and short papers)}, 4171--4186.

\bibitem[{Francis(2023)}]{francis2023personality}
Francis, S. E. X.~B. 2023.
\newblock \emph{The Personality of Depression-Utilizing Big Five Personality Traits to Detect Depression on Social Media}.
\newblock Master's thesis, NTNU.

\bibitem[{Fu et~al.(2025{\natexlab{a}})Fu, Fu, Zhang, Kuang, Dong, Su, Su, Shi, Yao, Zhao et~al.}]{fu2025first}
Fu, C.; Fu, Z.; Zhang, Q.; Kuang, X.; Dong, J.; Su, K.; Su, Y.; Shi, W.; Yao, J.; Zhao, Y.; et~al. 2025{\natexlab{a}}.
\newblock The First MPDD Challenge: Multimodal Personality-aware Depression Detection.
\newblock \emph{arXiv preprint arXiv:2505.10034}.

\bibitem[{Fu et~al.(2022)Fu, Liu, Ishi, and Ishiguro}]{fu2022adversarial}
Fu, C.; Liu, C.; Ishi, C.~T.; and Ishiguro, H. 2022.
\newblock An adversarial training based speech emotion classifier with isolated gaussian regularization.
\newblock \emph{IEEE Transactions on Affective Computing}, 14(3): 2361--2374.

\bibitem[{Fu et~al.(2025{\natexlab{b}})Fu, Qian, Su, Su, Wang, Shi, Liu, Liu, and Ishi}]{fu2025himul}
Fu, C.; Qian, F.; Su, K.; Su, Y.; Wang, Z.; Shi, J.; Liu, Z.; Liu, C.; and Ishi, C.~T. 2025{\natexlab{b}}.
\newblock HiMul-LGG: A hierarchical decision fusion-based local--global graph neural network for multimodal emotion recognition in conversation.
\newblock \emph{Neural Networks}, 181: 106764.

\bibitem[{Fu et~al.(2025{\natexlab{c}})Fu, Qian, Su, Su, Song, Niu, Shi, Liu, Liu, Ishi et~al.}]{fu2025facial}
Fu, C.; Qian, F.; Su, Y.; Su, K.; Song, S.; Niu, M.; Shi, J.; Liu, Z.; Liu, C.; Ishi, C.~T.; et~al. 2025{\natexlab{c}}.
\newblock Facial action units guided graph representation learning for multimodal depression detection.
\newblock \emph{Neurocomputing}, 619: 129106.

\bibitem[{Fu et~al.(2025{\natexlab{d}})Fu, Su, Su, Qian, Zhang, Liu, Song, Yang, Lv, Shan et~al.}]{fu2025m}
Fu, C.; Su, K.; Su, Y.; Qian, F.; Zhang, Y.; Liu, C.; Song, S.; Yang, L.; Lv, X.; Shan, P.; et~al. 2025{\natexlab{d}}.
\newblock M 3 ADD: A Novel Benchmark for Physiology Signal-based Automatic Depression Detection with Multimodal Multitask Multievent Framework.
\newblock In \emph{ICASSP 2025-2025 IEEE International Conference on Acoustics, Speech and Signal Processing (ICASSP)}, 1--5. IEEE.

\bibitem[{Ghosal et~al.(2019)Ghosal, Majumder, Poria, Chhaya, and Gelbukh}]{ghosal2019dialoguegcn}
Ghosal, D.; Majumder, N.; Poria, S.; Chhaya, N.; and Gelbukh, A. 2019.
\newblock Dialoguegcn: A graph convolutional neural network for emotion recognition in conversation.
\newblock \emph{arXiv preprint arXiv:1908.11540}.

\bibitem[{Hazarika, Zimmermann, and Poria(2020)}]{hazarika2020misa}
Hazarika, D.; Zimmermann, R.; and Poria, S. 2020.
\newblock Misa: Modality-invariant and-specific representations for multimodal sentiment analysis.
\newblock In \emph{Proceedings of the 28th ACM international conference on multimedia}, 1122--1131.

\bibitem[{He et~al.(2016)He, Zhang, Ren, and Sun}]{he2016deep}
He, K.; Zhang, X.; Ren, S.; and Sun, J. 2016.
\newblock Deep residual learning for image recognition.
\newblock In \emph{Proceedings of the IEEE conference on computer vision and pattern recognition}, 770--778.

\bibitem[{Kanchapogu and Mohanty(2025)}]{kanchapogu2025deep}
Kanchapogu, N.~R.; and Mohanty, S.~N. 2025.
\newblock Deep learning with ensemble-based hybrid AI model for bipolar and unipolar depression detection using demographic and behavioral based on time-series data.
\newblock \emph{Dialogues in Clinical Neuroscience}, 27(1): 16.

\bibitem[{Kazakos et~al.(2019)Kazakos, Nagrani, Zisserman, and Damen}]{kazakos2019epic}
Kazakos, E.; Nagrani, A.; Zisserman, A.; and Damen, D. 2019.
\newblock Epic-fusion: Audio-visual temporal binding for egocentric action recognition.
\newblock In \emph{Proceedings of the IEEE/CVF international conference on computer vision}, 5492--5501.

\bibitem[{Li et~al.(2024)Li, Dong, Yi, Liang, and Yan}]{li2024hypergraph}
Li, X.; Dong, Y.; Yi, Y.; Liang, Z.; and Yan, S. 2024.
\newblock Hypergraph Neural Network for Multimodal Depression Recognition.
\newblock \emph{Electronics}, 13(22): 4544.

\bibitem[{Locatello et~al.(2019)Locatello, Bauer, Lucic, Raetsch, Gelly, Sch{\"o}lkopf, and Bachem}]{locatello2019challenging}
Locatello, F.; Bauer, S.; Lucic, M.; Raetsch, G.; Gelly, S.; Sch{\"o}lkopf, B.; and Bachem, O. 2019.
\newblock Challenging common assumptions in the unsupervised learning of disentangled representations.
\newblock In \emph{international conference on machine learning}, 4114--4124. PMLR.

\bibitem[{Ma et~al.(2016)Ma, Yang, Chen, Huang, and Wang}]{ma2016depaudionet}
Ma, X.; Yang, H.; Chen, Q.; Huang, D.; and Wang, Y. 2016.
\newblock Depaudionet: An efficient deep model for audio based depression classification.
\newblock In \emph{Proceedings of the 6th international workshop on audio/visual emotion challenge}, 35--42.

\bibitem[{Meng et~al.(2024)Meng, Zhang, Shou, Shao, Ai, and Li}]{meng2024masked}
Meng, T.; Zhang, F.; Shou, Y.; Shao, H.; Ai, W.; and Li, K. 2024.
\newblock Masked graph learning with recurrent alignment for multimodal emotion recognition in conversation.
\newblock \emph{IEEE/ACM Transactions on Audio, Speech, and Language Processing}.

\bibitem[{Meng et~al.(2021)Meng, Speier, Ong, and Arnold}]{meng2021bidirectional}
Meng, Y.; Speier, W.; Ong, M.~K.; and Arnold, C.~W. 2021.
\newblock Bidirectional representation learning from transformers using multimodal electronic health record data to predict depression.
\newblock \emph{IEEE journal of biomedical and health informatics}, 25(8): 3121--3129.

\bibitem[{M{\o}ller et~al.(2023)M{\o}ller, Dalsgaard, Pera, and Aiello}]{moller2023prompt}
M{\o}ller, A.~G.; Dalsgaard, J.~A.; Pera, A.; and Aiello, L.~M. 2023.
\newblock Is a prompt and a few samples all you need? Using GPT-4 for data augmentation in low-resource classification tasks.
\newblock \emph{arXiv preprint arXiv:2304.13861}, 4(13861): 1--12.

\bibitem[{Niu, Li, and Fu(2024)}]{niu2024pointtransform}
Niu, M.; Li, M.; and Fu, C. 2024.
\newblock Pointtransform networks for automatic depression level prediction via facial keypoints.
\newblock \emph{Knowledge-Based Systems}, 297: 111951.

\bibitem[{Pan et~al.(2024)Pan, Shang, Liu, Shao, Guo, Ding, and Hu}]{pan2024spatial}
Pan, Y.; Shang, Y.; Liu, T.; Shao, Z.; Guo, G.; Ding, H.; and Hu, Q. 2024.
\newblock Spatial--temporal attention network for depression recognition from facial videos.
\newblock \emph{Expert systems with applications}, 237: 121410.

\bibitem[{Qin et~al.(2022)Qin, Lei, Pinaya, Pan, Li, Zhu, Sweeney, Mechelli, and Gong}]{qin2022using}
Qin, K.; Lei, D.; Pinaya, W.~H.; Pan, N.; Li, W.; Zhu, Z.; Sweeney, J.~A.; Mechelli, A.; and Gong, Q. 2022.
\newblock Using graph convolutional network to characterize individuals with major depressive disorder across multiple imaging sites.
\newblock \emph{EBioMedicine}, 78.

\bibitem[{Ravi et~al.(2024)Ravi, Wang, Flint, and Alwan}]{ravi2024enhancing}
Ravi, V.; Wang, J.; Flint, J.; and Alwan, A. 2024.
\newblock Enhancing accuracy and privacy in speech-based depression detection through speaker disentanglement.
\newblock \emph{Computer speech \& language}, 86: 101605.

\bibitem[{Rohanian et~al.(2019)Rohanian, Hough, Purver et~al.}]{rohanian2019detecting}
Rohanian, M.; Hough, J.; Purver, M.; et~al. 2019.
\newblock Detecting Depression with Word-Level Multimodal Fusion.
\newblock In \emph{Interspeech}, 1443--1447.

\bibitem[{Shen, Yang, and Lin(2022)}]{shen2022automatic}
Shen, Y.; Yang, H.; and Lin, L. 2022.
\newblock Automatic depression detection: An emotional audio-textual corpus and a gru/bilstm-based model.
\newblock In \emph{ICASSP 2022-2022 IEEE International Conference on Acoustics, Speech and Signal Processing (ICASSP)}, 6247--6251. IEEE.

\bibitem[{Sun et~al.(2023)Sun, Liu, Chen, and Lin}]{sun2023modality}
Sun, H.; Liu, J.; Chen, Y.-W.; and Lin, L. 2023.
\newblock Modality-invariant temporal representation learning for multimodal sentiment classification.
\newblock \emph{Information Fusion}, 91: 504--514.

\bibitem[{Tan et~al.(2025)Tan, Kwan, Ng, and Hum}]{tan2025psy}
Tan, J.~J.; Kwan, B.-H.; Ng, D.; and Hum, Y. 2025.
\newblock Psychology-informed Natural Language Understanding: Integrating Personality and Emotion-aware Features for Comprehensive Sentiment Analysis and Depression Detection.
\newblock \emph{Pertanika Journal of Science and Technology}, 33.

\bibitem[{Tian et~al.(2024)Tian, Anantrasirichai, Nicholson, and Achim}]{tian2024tagat}
Tian, X.; Anantrasirichai, N.; Nicholson, L.; and Achim, A. 2024.
\newblock TaGAT: Topology-Aware Graph Attention Network for Multi-modal Retinal Image Fusion.
\newblock In \emph{International Conference on Medical Image Computing and Computer-Assisted Intervention}, 775--784. Springer.

\bibitem[{Trotzek, Koitka, and Friedrich(2018)}]{trotzek2018utilizing}
Trotzek, M.; Koitka, S.; and Friedrich, C.~M. 2018.
\newblock Utilizing neural networks and linguistic metadata for early detection of depression indications in text sequences.
\newblock \emph{IEEE Transactions on Knowledge and Data Engineering}, 32(3): 588--601.

\bibitem[{Wang, Ravi, and Alwan(2023)}]{wang2023non}
Wang, J.; Ravi, V.; and Alwan, A. 2023.
\newblock Non-uniform speaker disentanglement for depression detection from raw speech signals.
\newblock In \emph{Interspeech}, volume 2023, 2343.

\bibitem[{Wang et~al.(2020)Wang, Zhu, Bo, Cui, Shi, and Pei}]{wang2020gcn}
Wang, X.; Zhu, M.; Bo, D.; Cui, P.; Shi, C.; and Pei, J. 2020.
\newblock Am-gcn: Adaptive multi-channel graph convolutional networks.
\newblock In \emph{Proceedings of the 26th ACM SIGKDD International conference on knowledge discovery \& data mining}, 1243--1253.

\bibitem[{{WHO}(2023)}]{WHO}
{WHO}. 2023.
\newblock Depressive disorder (depression).
\newblock \url{https://www.who.int/news-room/fact-sheets/detail/depression}.
\newblock Accessed: 2025-06-05.

\bibitem[{Wu et~al.(2024)Wu, Wang, Wang, and Zheng}]{wu2024large}
Wu, Y.; Wang, Y.; Wang, C.; and Zheng, Z. 2024.
\newblock Large Language Model Enhanced Machine Learning Estimators for Classification.
\newblock \emph{arXiv preprint arXiv:2405.05445}.

\bibitem[{Wu et~al.(2025)Wu, Gong, Ai, Shi, Donbekci, and Hirschberg}]{wu2025beyond}
Wu, Z.; Gong, Z.; Ai, L.; Shi, P.; Donbekci, K.; and Hirschberg, J. 2025.
\newblock Beyond Silent Letters: Amplifying {LLM}s in Emotion Recognition with Vocal Nuances.
\newblock In Chiruzzo, L.; Ritter, A.; and Wang, L., eds., \emph{Findings of the Association for Computational Linguistics: NAACL 2025}, 2202--2218. Albuquerque, New Mexico: Association for Computational Linguistics.
\newblock ISBN 979-8-89176-195-7.

\bibitem[{Ye et~al.(2024{\natexlab{a}})Ye, Yu, Lu, Wang, Zheng, Liu, and Wang}]{ye2024dep}
Ye, J.; Yu, Y.; Lu, L.; Wang, H.; Zheng, Y.; Liu, Y.; and Wang, Q. 2024{\natexlab{a}}.
\newblock DEP-Former: Multimodal Depression Recognition Based on Facial Expressions and Audio Features via Emotional Changes.
\newblock \emph{IEEE Transactions on Circuits and Systems for Video Technology}.

\bibitem[{Ye, Zhang, and Shan(2025)}]{ye2025depmamba}
Ye, J.; Zhang, J.; and Shan, H. 2025.
\newblock Depmamba: Progressive fusion mamba for multimodal depression detection.
\newblock In \emph{ICASSP 2025-2025 IEEE International Conference on Acoustics, Speech and Signal Processing (ICASSP)}, 1--5. IEEE.

\bibitem[{Yi et~al.(2024)Yi, Zhao, Shen, and Zhang}]{yi2024multimodal}
Yi, Z.; Zhao, Z.; Shen, Z.; and Zhang, T. 2024.
\newblock Multimodal Fusion via Hypergraph Autoencoder and Contrastive Learning for Emotion Recognition in Conversation.
\newblock In \emph{Proceedings of the 32nd ACM International Conference on Multimedia}, 4341--4348.

\bibitem[{Yu et~al.(2024)Yu, Wang, Wang, Luo, and Zhou}]{yu2024towards}
Yu, T.; Wang, J.; Wang, J.; Luo, J.; and Zhou, G. 2024.
\newblock Towards Emotion-enriched Text-to-Motion Generation via LLM-guided Limb-level Emotion Manipulating.
\newblock In \emph{Proceedings of the 32nd ACM International Conference on Multimedia}, 612--621.

\bibitem[{Zellinger et~al.(2019)Zellinger, Moser, Grubinger, Lughofer, Natschl{\"a}ger, and Saminger-Platz}]{zellinger2019robust}
Zellinger, W.; Moser, B.~A.; Grubinger, T.; Lughofer, E.; Natschl{\"a}ger, T.; and Saminger-Platz, S. 2019.
\newblock Robust unsupervised domain adaptation for neural networks via moment alignment.
\newblock \emph{Information Sciences}, 483: 174--191.

\bibitem[{Zhang et~al.(2023)Zhang, Li, Zhang, Ng, Leau, and Yan}]{zhang2023deep}
Zhang, L.-M.; Li, Y.; Zhang, Y.-T.; Ng, G.~W.; Leau, Y.-B.; and Yan, H. 2023.
\newblock A deep learning method using gender-specific features for emotion recognition.
\newblock \emph{Sensors}, 23(3): 1355.

\bibitem[{Zhao et~al.(2018)Zhao, Ding, Han, and Gao}]{zhao2018personality}
Zhao, S.; Ding, G.; Han, J.; and Gao, Y. 2018.
\newblock Personality-Aware Personalized Emotion Recognition from Physiological Signals.
\newblock In \emph{IJCAI}, 1660--1667.

\bibitem[{Zhao et~al.(2025{\natexlab{a}})Zhao, Zhang, Su, Su, Liu, Wang, and Yu}]{zhao2025depressionmignn}
Zhao, S.; Zhang, Y.; Su, Y.; Su, K.; Liu, J.; Wang, T.; and Yu, S. 2025{\natexlab{a}}.
\newblock DepressionMIGNN: A Multiple-Instance Learning-Based Depression Detection Model with Graph Neural Networks.
\newblock \emph{Sensors}, 25(14): 4520.

\bibitem[{Zhao et~al.(2025{\natexlab{b}})Zhao, Zhang, Li, Song, Lian, Liu, Wang, and Fu}]{zhao2025chinese}
Zhao, Y.; Zhang, H.; Li, J.; Song, S.; Lian, C.; Liu, Y.; Wang, Y.; and Fu, C. 2025{\natexlab{b}}.
\newblock A Chinese multimodal depression dataset with personality labeling for older adults with underlying medical conditions.
\newblock \emph{IEEE Transactions on Affective Computing}.

\end{thebibliography}

\end{document}